\documentclass[10pt,twocolumn,letterpaper]{article}
\usepackage{titlesec}
\titlespacing*{\paragraph}
{0pt}{\smallskipamount}{\smallskipamount}
\usepackage{cvpr}              % To produce the CAMERA-READY version
\usepackage{algorithm}
\usepackage{algorithmic}
\usepackage{multirow}
\usepackage{amsfonts}
\usepackage{amsmath}

\newcommand{\boldZ}{\boldsymbol{Z}}

\newcommand{\frameworkName}{GKDE\xspace}
\newcommand{\std}[1]{\footnotesize{#1}}

\newlength\savewidth\newcommand\shline{\noalign{\global\savewidth\arrayrulewidth\global\arrayrulewidth1.25pt}\hline\noalign{\global\arrayrulewidth\savewidth}}
\usepackage[dvipsnames]{xcolor}

\definecolor{cvprblue}{rgb}{0.21,0.49,0.74}
\usepackage[pagebackref,breaklinks,colorlinks,citecolor=cvprblue]{hyperref}

\def\paperID{} % *** Enter the Paper ID here
\def\confName{Arxiv}
\def\confYear{2024}

\title{Density Distribution-based Learning Framework for Addressing Online Continual Learning Challenges}

\author{Shilin Zhang\\
North China University of Technology\\
Shijingshan District Beijing\\
{\tt\small zhangshilin@ncut.edu.cn}
\and
Jiahui Wang\\
North China University of Technology\\
Shijingshan District Beijing\\
{\tt\small secondauthor@i2.org}
}

\begin{document}
\maketitle
\begin{abstract}
In this paper, we address the challenges of online Continual Learning (CL) by introducing a density distribution-based learning framework. CL, especially the Class Incremental Learning, enables adaptation to new test distributions while continuously learning from a single-pass training data stream, which is more in line with the practical application requirements of real-world scenarios. However, existing CL methods often suffer from catastrophic forgetting and higher computing costs due to complex algorithm designs, limiting their practical use. Our proposed framework overcomes these limitations by achieving superior average accuracy and time-space efficiency, bridging the performance gap between CL and classical machine learning. Specifically, we adopt an independent Generative Kernel Density Estimation (GKDE) model for each CL task. During the testing stage, the GKDEs utilize a self-reported max probability density value to determine which one is responsible for predicting incoming test instances. A GKDE-based learning objective can ensure that samples with the same label are grouped together, while dissimilar instances are pushed farther apart. Extensive experiments conducted on multiple CL datasets validate the effectiveness of our proposed framework. Our method outperforms popular CL approaches by a significant margin, while maintaining competitive time-space efficiency, making our framework suitable for real-world applications. Code will be available at https://github.com/xxxx/xxxx.
\end{abstract}    
\section{Introduction}
\label{intro}
Despite the impressive performance of current artificial intelligence systems \cite{ResNet, ViT} on various tasks, their performance can significantly degrade when tested on unseen data in new environments. This phenomenon, known as \emph{catastrophic forgetting} (CF) \cite{catastrophic, catastrophic2}, occurs when previously learned knowledge is forgotten while learning new tasks in Continual Learning (CL) settings \cite{survey1, survey2, survey3, CL_CIL1}. Online CL aims to continuously learn from a non-stationary data stream, adapting to new instances and mimicking human general intelligence. In this paper, we specifically focus on Class Incremental Learning (CIL) in the online CL mode \cite{online_survey, onlineCL1, onlineCL2, ASER}. In CIL, the model incrementally learns labels in a sequence of tasks from a single-pass data stream and does not have access to task information during inference.

Existing online CL models \cite{ASER, SCR, DVC, online_pro_accum, ER, ER_AML, onlineCL1} face several challenges that hinder their practical application. Replay-based methods \cite{ER, SCR, OCM, MIR, DVC} store a portion of training samples to facilitate learning of upcoming tasks but require additional memory usage. Parameter-isolation methods such as HAT \cite{Serra2018overcoming} and SupSup \cite{supsup2020} train and protect a sub-network for each task but suffer from complex network designs and low training efficiency \cite{Mallya2017packnet, abati2020conditional, von2019continual_hypernet, rajasegaran2020itaml, hung2019compact, henning2021posterior}. Moreover, there is a substantial performance gap between popular CL methods and classic machine learning (ML) algorithms, and current feature representations fail to capture causal factors for making predictions, resulting in limited generalization ability.

To address these challenges, we propose a novel density distribution-based learning framework called Generative Kernel Density Estimation (GKDE) for online CL. Our method replaces the traditional cross-entropy loss with a probability density-based learning objective and builds upon elementary Residual or Transformer Networks \cite{ResNet, ViT} with a simple linear projection head. The GKDE framework is both time and space efficient due to its simple network structure and straightforward training procedure. In addition, our method trains an independent model for each task considering the diverse and evolving nature of future tasks while addressing the limitations of existing CL models. During inference, all models in the Model Bank (MB) report their probability density values for a given test sample, allowing us to determine which model is responsible for the current prediction. This mechanism enables our model to achieve state-of-the-art average accuracy, effectively bridging the performance gap between CL and classical ML. Furthermore, our method's feature representation aligns with probability density values, enhancing the capture of causal factors while discouraging the influence of spurious features, which leads to improved model generalization. Extensive experiments conducted on three CL datasets validate the effectiveness of our proposed method.

The main contributions of this paper are:

\begin{enumerate}[leftmargin=12pt, itemsep=0pt, topsep=0pt, partopsep=0pt, noitemsep]
	\item We propose a novel density-based learning framework called Generative Kernel Density Estimation (GKDE) designed to address the inherent difficulties of online continual learning by improving feature representations. Our method advances the state-of-the-art by effectively eliminating CF and improving model generalization in evolving environments.
	\item The GKDE framework is both time and space efficient due to its simple network structure and straightforward training procedure, and compared with other methods, the time-space efficiency has obvious advantages, making it well-suited for practical deployment in online learning scenarios.
	\item Through the feature representation aligned with probability density values, our method demonstrates enhanced model generalization, effectively addressing the challenges of continual learning in dynamic environments. This improvement is verified through experiments in three CL datasets with superior average accuracy, for instance our method based on VIT achieves  99.25\% on CIFAR-10 and 99.10\% on TinyImageNet.
\end{enumerate}
\section{Related Works}
\textit{Parameter-isolation methods} aim to protect specific network parameters for each task during the training stage \cite{Mallya2017packnet,abati2020conditional}. These methods, such as HAT \cite{Serra2018overcoming} and SupSup \cite{supsup2020}, effectively relieve the issue of CF. In contrast, other approaches \cite{PNN, para-iso1, pathNet} tackle the forgetting problem by dynamically allocating parameters or modifying the network's architecture. However, a limitation of these methods is that they require task-ids to be provided for test instances and the training is not efficient due to complicated network structure. Distinguishing itself from the aforementioned methods, our proposed GKDE learning framework achieves task-id identification (TP) and within-task prediction (WP) simultaneously. This is accomplished by employing a simple baseline network and a projection head, which transform the CL into a classical ML.

\textit{Replay-based methods}~\cite{ER, MIR, GSS, AGEM, DER++, GDumb} have been proposed to mitigate the problem of CF \cite{ER, MIR, GSS, AGEM, DER++, GDumb}. These methods typically maintain a memory buffer that stores samples from previous tasks. For instance, Experience Replay \cite{ER} randomly samples from the buffer to provide additional training data. In contrast, MIR \cite{MIR} retrieves instances from a memory bank by comparing the interference of losses. In the online setting, ASER \cite{ASER} introduces a buffer management algorithm that optimizes the sampling process. Additionally, OCM \cite{OCM} prevents forgetting by maximizing mutual information. Diverging from these approaches, our work takes a different perspective on CF. We propose a novel probability density learning framework that eliminates the need to store training samples. Instead, we focus on learning representative and discriminative features through distribution density estimation. This approach is memory-efficient, as it only requires low-level randomly generated training features for each task. By rethinking the problem from this new perspective, we aim to provide an innovative solution to address CF issue in CL.
\section{Proposed Method}
We start by explaining our overall architecture. Then, we describe the definition of the GKDE, the training objective, and the theoretical analysis of the method successively. 

\subsection{Overall Architecture}
\begin{figure*}[htp]
  \centering
  \includegraphics[width=0.9\linewidth]{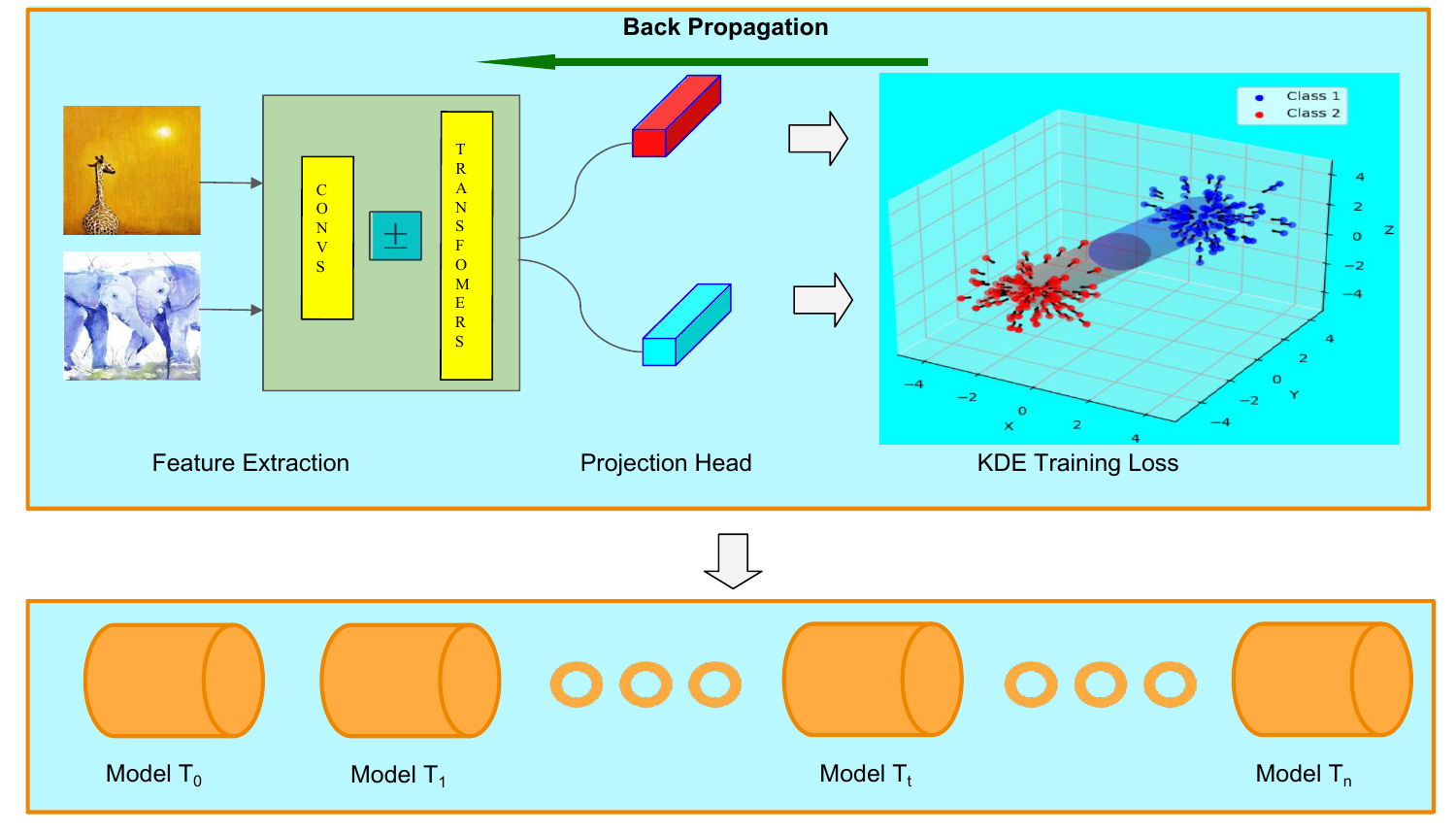}
  \caption{Illustration of the proposed simple yet effective GKDE framework for online CL. In the upper left of the figure, Giraffe and Elephant are represented by two PDFs in the embedding feature space. The training objective of our proposed KDE is to bring samples with the same label closer in higher probability regions and push dissimilar samples, as shown in the top right of the figure, away from this confusing region. During inference, the probability of a test sample is computed against each PDF in the Model Bank. The selected model is determined based on the highest probability density value, allowing for WP.
  }
  \label{fig:framework}
\end{figure*}

Fig.~\ref{fig:framework} presents the illustration of the proposed framework. The backbone network, such as ResNet or Transformer, serves as the feature extraction component without any modifications. A projection head is employed to generate $d$-dimensional probability density values, which replaces the traditional linear classification layer and also the cross entropy loss. In our approach, the class in each task, such as Giraffe or Elephant in Fig.~\ref{fig:framework}, is represented by a probability density function (PDF). The GKDE training objective is designed to bring samples with the same label closer together in higher probability regions, while pushing dissimilar samples, shown in the top right of Fig.~\ref{fig:framework}, further away from this ambiguous region.
During the inference stage, the probability of a test sample is computed against each PDF in the Model Bank (MB). By selecting the PDF with the highest probability, the correct model is self recommended in the TP procedure, enabling high-performance WP.

In the inference stage for a test instance without task ID, the prediction probability is the product of two produced probabilities: \textit{WP} and \textit{TP}, 
\begin{align}
	\label{tp_and_wp}
	\mathbf{P}(X_{k,j} | x)
	&= \mathbf{P} (X_{k,j} | x, k) \mathbf{P}(X_{k} | x), 
\end{align}
where $X_{k,j}$ represents instances belonging to task $k$ and class label $j$. When it comes to a test instance $x$, we compute two probabilities using the same GKDE framework. The first probability, located on the right-hand side (RHS), represents WP. The second probability, represents TP. It is worth noting that our method stands out from others in that both probabilities are computed using a single GKDE framework. In contrast, alternative methods often rely on separate models or subnetworks to calculate these probabilities. By unifying the computation within a single framework, our approach offers a more streamlined and efficient solution. 

We outline the primary steps of our method as follows:

\begin{itemize}
	\item \textbf{Initial PDF Generation:} Utilizing a pre-trained backbone, the initial Probability Density Function (PDF) is generated, corresponding to the embedding features of the training samples. This initial stage is illustrated in the top-left section of Fig.~\ref{fig:framework}.
	
	\item \textbf{Feature Embeddings Optimization:} Our proposed GKDE operates as the learning objective. It optimizes the feature extraction process, amplifying casual factors, attenuating spurious ones and thus attaining general feature representations in alignment with each class PDF. This is depicted in the top-right section of Fig.~\ref{fig:framework}.
	
	\item \textbf{Model Bank Management:} Our Model Bank houses the trained networks and PDFs for all tasks. During the inference stage, these elements are applied to accomplish TP and WP classifications simultaneously. This is demonstrated in the bottom section of Fig.~\ref{fig:framework}.
\end{itemize}

\subsection{Generative Kernel Density Estimation}
In this paper, we consider the dataset distribution represented as $(\boldsymbol{X}, Y)$, where $\boldsymbol{X}$ represents images, $\boldsymbol{Z}$ the features, and $Y$ the ground truth labels. $\boldZ = (Z_1, Z_2, \dots Z_d)^T \in \mathbb{R}^d$ is the $d$-dimensional features extracted by backbones and $Y \in \mathbb{R}$ is the target. In contrast to other methods, we use the GKDE to describe each class of a specific task in the feature embedding space to align $\boldsymbol{Z}$ with PDFs and measure the distribution shifts between the test sample and each PDF in our MB. By aggregating instances of each category into a compact PDF, we can effectively measure and predict samples in the unknown test distribution, achieving both TP and WP simultaneously. 

Formally, online CIL processes a continuous sequence of tasks from a single-pass data stream $\mathfrak{D}=\left\{\mathcal{D}_1, \ldots, \mathcal{D}_T \right\} $, where $\mathcal{D}_t = \left\{ x_{i}, y_{i} \right\} ^{N_t}_{i=1} $ is the training set of task $t$, and $T$ is the overall number of tasks. Training Set $\mathcal{D}_t$ contains $N_t$ samples, $y_{i}$ is the label of sample $x_{i}$ and $y_{i} \in \mathcal{C}_t$, where $\mathcal{C}_t$ is the label set of task $t$ and the labels in each task are different. The objective of online CIL is to achieve accurate predictions for incoming instances, regardless of whether they belong to an older or current task.

In this paper, $\phi_\theta(\cdot)$ represents the feature extractor parameterized by $\theta$, and $f_\psi(\cdot)$ represents the projection component for the GKDE framework, where $\psi$ is the parameter to be learned. The features generated by $\phi_\theta(\cdot)$ are further transformed into a $d$-dimensional feature space by the projection head denoted as $\boldZ$. For training samples in the embedding space $\mathbf{z}_1,\ldots,\mathbf{z}_n$ in $\mathbb{R}^d$, the GKDE is evaluated at $\mathbf{x} \in \mathbb{R}^d$:
\begin{align}
	\label{general}
	\hat{k}(\mathbf{z};\mathbf{H}):=\frac{1}{n|\mathbf{H}|^{1/2}}\sum_{i=1}^nK\left(\mathbf{H}^{-1/2}(\mathbf{z}-\mathbf{z}_i)\right),
\end{align}
where $K$ is a multi-variate kernel and $\mathbf{H}$ is a bandwidth matrix, which is $d \times d$ symmetric and positive definite. For the sake of simplicity, we consider a Gaussian function as the kernel $K$ in our training objective, where $\boldsymbol{\mu}$ and $\boldsymbol{\Sigma}$ are computed from the samples of a specific class in a task. The bandwidth matrix $\mathbf{H}$ is replaced by a scalar value $h$, which is applied to each dimension. As a result, the GKDE can be further simplified as: 
\begin{equation}
	\label{specific}
	\hat{k}(\mathbf{z},h) = \frac{1}{nh^d} \sum_{i=1}^{n} K(\frac{\mathbf{z} - {\mathbf{z}}_i}{h}) = \frac{1}{n} \sum_{i=1}^{n} K_h (\mathbf{z} - {\mathbf{z}}_i)\,
\end{equation}
In Eq.~\ref{general} and Eq.~\ref{specific}, $\mathbf{z}$ represents the embedding feature of a training sample. On the other hand, $\mathbf{z_i}$ serves as the anchor point for a specific PDF to facilitate the computation of GKDE. These anchor points are randomly generated from the original training features and the bandwidth is used as the standard deviation to provide some randomness. We sample $n$ feature embeddings for each class of a specific task to act as anchor points. Therefore, unlike replay-based CL, our framework does not require storing any training samples. Instead, feature embeddings are generated based on the training samples, incorporating some uncertainties. This feature randomness is beneficial for enhancing model generalization, as confirmed by \cite{li2022uncertainty}. The specific value of the sample size $n$ will be discussed in the experimental section.

\subsection{The Training Objective Design}
In this part, we design our learning objective to serve as the training loss to replace the cross entropy and guide the network training process. Let’s denote by $k^{j=y}$ as the conditional PDF of $\mathbf{Z}|Y=j$. $\pi_j$ is the value of the probability mass function of  Y at $j$, i.e., $\mathbb{P}[Y=j]=\pi_j$, for $j=1,\ldots, m$, and $m$ is the number of distinct labels in the training set of a specific task. Estimating the class probabilities $\pi_j$ is based on their relative frequencies $\hat{\pi}_j:=\frac{n_j}{n}$, where $n_j:=\#\{i=1,\ldots,n:Y_i=j\}$. 
Therefore, the learning objective for the $j_{th}$ class is designed as:
\begin{align}
	\label{loss}
	\mathcal{L}(\mathbf{z})=-\hat{\pi}_j\hat{k}^{j = y}(\mathbf{z})+ \sum_{j=1,j \neq y}^m \hat{\pi}_j\hat{k}^{j \neq y}(\mathbf{z}),
\end{align}
where $\mathbf{z}$ is the embedding feature of a training sample and $\mathbf{y}$ is its corresponding label. In Eq.~\ref{loss}, there are two components and the first one aims to maximize the probability as the sample $\mathbf{z}$ belongs to the PDF $\hat{k}^{j = y}(\mathbf{z})$. While for the second term, we should minimize the probability as the training feature $\mathbf{z}$ has nothing to do with the PDF $\hat{k}^{j \neq y}(\mathbf{z})$. Guided by this learning objective, samples with the same label are encouraged to be grouped together in a region of higher probability, while dissimilar samples are pushed away to a region of marginal probability. Additionally, considering the training feature $\mathbf{z}$, the loss of the GKDE can be expressed in terms of the learnable parameters $\mathbf{\theta}$ and $\mathbf{\psi}$.

\begin{align}
	\label{loss_final}
	\begin{aligned}
		\mathcal{L}(\mathbf{\theta},\mathbf{\psi}; \mathbf{z})&= -\hat{\pi}_j \sum_{i=1}^{n} K^{j = y}_h (\mathbf{z} - {\mathbf{z}}_i)\\
		&+ \hat{\pi}_j \sum_{j=1,j \neq y}^m \sum_{i=1}^{n} K^{j \neq y}_h (\mathbf{z} - {\mathbf{z}}_i)\\	
		&= -\hat{\pi}_j \sum_{i=1}^{n} K^{j = y}_h (\mathbf{f}_{\psi}(\mathbf{\phi}_{\theta}(\mathbf{x})) - \mathbf{z}_i)\\
		& +\hat{\pi}_j \sum_{j=1,j \neq y}^m \sum_{i=1}^{n} K^{j \neq y}_h (\mathbf{f}_{\psi}(\mathbf{\phi}_{\theta}(\mathbf{x})) - \mathbf{z}_i).
	\end{aligned}
\end{align}

In Eq.~\ref{loss_final}, $\mathbf{\phi}_{\theta}(\cdot)$ and $\mathbf{f}_{\psi}(\cdot)$ represent the feature extractor and the projection head, respectively. These can be implemented using models such as ResNet \cite{stable_learning} or Transformer \cite{transformer}, and they contain the trainable parameters $\theta$ and $\psi$. Here, $\mathbf{x}$ denotes the training sample and $\mathbf{z_i}$ represents the anchor feature in the $j_{th}$ PDF, where $j$ is $\mathbf{x}$'s ground truth label. Eq.~\ref{loss_final} is utilized as a loss function in our framework. Since the probability values involved in the loss computation for each anchor are in logarithmic format, we apply a clipping operation to ensure the values are not less than a threshold $\gamma$. This clipping operation not only stabilizes the training process but also prevents the values from underflow during backpropagation.

In the testing stage, for each test instance, the task label is determined through the TP procedure.
\begin{align}
	\label{inference}
	\begin{aligned}
		\mathcal{T}= \mathop{\arg\max}\limits_{t} \mathop{\arg\max}\limits_{j} k_{j}^t(\mathbf{f}_{\psi}^t(\mathbf{\phi}_{\theta}^t(\mathbf{x}))),
	\end{aligned}
\end{align}
where $t$ represents the task index that has been previously trained and stored in the MB. $j$ represents the class label within task $t$, and $\mathcal{T}$ denotes the predicted task ID.

For WP, to assign an observation $\mathbf{z}$ of $\mathbf{Z}$
to each of the classes of $\mathbf{Y}$ can be solved by maximizing the conditional probability:
\begin{align}
	\mathbb{P}[Y=j|\mathbf{Z}=\mathbf{z}]&=\frac{k_j^{\mathcal{T}}(\mathbf{z})\mathbb{P}[Y=j]}{k^{\mathcal{T}}(\mathbf{z})}\nonumber\\
	&=\frac{\pi_jk_j^{\mathcal{T}}(\mathbf{z})}{\sum_{j=1}^m \pi_jk_j^{\mathcal{T}}(\mathbf{z})},
\end{align}
where $\mathbf{z}$ is assigned to the most likely class $j$. 
\begin{align}
	\label{WP}
	k_{\mathrm{Bayes}}^{\mathcal{T}}(\mathbf{z}):=\arg\max_{j=1,\ldots,m} \pi_jk_j^{\mathcal{T}}(\mathbf{z}).	
\end{align}

According to Eq.~\ref{inference}, we can achieve TP and select the correct model corresponding to $\mathcal{T}$. Subsequently, based on Eq.~\ref{WP}, we can achieve the WP. Therefore, the procedure defined in Eq.~\ref{tp_and_wp} for a test instance is implemented in an online CL fashion.

\subsection{Theoretical Analysis of Model Generalization}

The GKDE framework exhibits excellent generality even though the test instances have severe co-variate shift. In this part, we give some theoretical analysis of the model generality. In the training process, the derivatives of the GKDE loss w.r.t parameters $\theta$ and $\psi$ are updated in the backward propagation. For a PDF $k(\cdot):\mathbb{R}^d\longrightarrow\mathbb{R}$, its gradient $\mathrm{D}k(\mathbf{z}):\mathbb{R}^d\longrightarrow\mathbb{R}^d$ is defined as:

\begin{align}
	\mathrm{D}k(\mathbf{z}):=\begin{pmatrix}\frac{\partial k(\mathbf{z})}{\partial \mathbf{z_1}}\\\vdots\\\frac{\partial k(\mathbf{z})}{\partial \mathbf{z_d}}\end{pmatrix}.	
\end{align}

The Hessian of $k(\cdot)$ is:
\begin{align}
	\mathrm{H}k(\mathbf{z}):=\left(\frac{\partial^2 k(\mathbf{z})}{\partial \mathbf{z_i}\partial \mathbf{z_j}}\right),\quad i,j=1,\ldots,d.
\end{align}

To quantify the bias and variance of the GKDE, which is commonly used to assess the generalizability of ML models, we employ the $\mathrm{vec}$ operator to concatenate the columns of the Hessian matrix into a single vector for easy expression, similar to the Kronecker product $\mathrm{D}^{\otimes 2}:=\mathrm{D}\otimes\mathrm{D}$. By doing so, we obtain Eq.~\ref{bias_and_variance}, which provides a characterization of the bias and variance of the GKDE at the point $\mathbf{z}$.

\begin{align}
	\label{bias_and_variance}
	\begin{aligned}
		\mathrm{Bias}[\hat{k}(\mathbf{z};\mathbf{H})]&=\frac{1}{2}\mu_2(K)(\mathrm{D}^{\otimes2}k(\mathbf{z}))'\mathrm{vec}\,\mathbf{H}+o(\|\mathrm{vec}\,\mathbf{H}\|),\\
		\mathbb{V}\mathrm{ar}[\hat{k}(\mathbf{z};\mathbf{H})]&=\frac{R(K)}{n|\mathbf{H}|^{1/2}}k(\mathbf{z})+o((n|\mathbf{H}|^{1/2})^{-1}),
	\end{aligned}
\end{align}
where $R(K)=\int K(\mathbf{z})^2\,\mathrm{d}\mathbf{z}$ and $\mu_2(K):=\int \mathbf{z}^2 K(\mathbf{z})\,\mathrm{d}\mathbf{z}$. The detailed computation can be found in the Appendix A. 

Eq.~\ref{bias_and_variance} presents a theoretical framework for comprehending how the GKDE achieves model generalization. The bias term in Eq.~\ref{bias_and_variance} represents the training error, while the variance term reflects the test error resulting from model overfitting. As the value of $\mathbf{H}$ decreases, the bias decreases as well, indicating a reduction in training errors. On the other hand, the variance serves as an indicator of overfitting. A high variance value suggests a decrease in model generalization, which is contingent on the value of $k(\mathbf{z})$. The variance decreases as a factor of $(n|\mathbf{H}|^{1/2})^{-1}$. Importantly, in higher dimensions, where $k(\mathbf{z})$ tends to be small, overfitting can be alleviated. The choice of bandwidth $\mathbf{H}$ and embedding dimension plays a crucial role in striking a balance between bias and variance. This delicate balance will be further investigated in our experimental settings.

\section{Experiments}
\subsection{Experimental Setup}
\paragraph{Datasets.}
We conducted our experiments using three image classification benchmark datasets: \textbf{CIFAR-10}~\cite{cifar10_100}, \textbf{CIFAR-100}~\cite{cifar10_100}, and \textbf{TinyImageNet}~\cite{tinyImageNet}, in order to evaluate the performance of online CIL methods. For CIFAR-10, following the approach of~\cite{ASER, SCR, DVC}, we divided the dataset into 5 disjoint tasks. Each task consisted of 2 disjoint classes, with 10,000 samples for training and 2,000 samples for testing. Similarly, for CIFAR-100, we split the dataset into 10 disjoint tasks, with each task containing 10 disjoint classes. The training set for each task comprised 5,000 samples, while the testing set had 1,000 samples. Regarding TinyImageNet, we followed the method outlined in~\cite{OCM} and split it into 100 disjoint tasks. Each task consisted of 2 disjoint classes, with 1,000 samples allocated for training and 100 samples for testing. 
\paragraph{Baselines and Evaluation metrics.}
We compare our framework with some online CL baselines: {AGEM}~\cite{AGEM}, {MIR}~\cite{MIR}, {GSS}~\cite{GSS}, {ER}~\cite{ER}, {GDumb}~\cite{GDumb}, {ASER}~\cite{ASER}, {SCR}~\cite{SCR}, {CoPE}~\cite{online_pro_ema}, {DVC}~\cite{DVC}, OnPro~\cite{onpro}, {OCM}~\cite{OCM} etc. We use Average Accuracy~\cite{ASER, DVC} to measure the performance of each online CIL method, which evaluates the accuracy of the test sets from all seen tasks, defined as $\text {Average Accuracy} =\frac{1}{T} \sum_{j=1}^T a_{T, j},$
where $a_{i, j}$ is the accuracy on task $j$ after the model is trained from task $1$ to $i$. 
Average Forgetting is not used in this paper, as the GKDE do not suffer from the DF issue, and it is almost approaching to zero.
\paragraph{Implementation details.}
We use ResNet18 \cite{ResNet} and VIT \cite{ViT} as the backbone $\phi_{\theta}$ and a linear layer as the projection head $f_{\psi}$, following the approach in SCR \cite{SCR}, OCM \cite{OCM}, and Co2L \cite{Co2L}. The hidden dimension $d$ is a hyperparameter, which will be incrementally increased from 2 to 128 and verified in the experiments. We do not employ a classification layer or entropy loss in this paper. For training the model, we use the Adam optimizer with an initial learning rate of $5\times10^{-4}$ for all datasets. The weight decay is set to $1.0\times10^{-4}$. The batch size $N$ is set to 128. To prevent underflow, log probabilities are clipped when they fall below -700. We report the average results across 15 runs for all experiments.

\subsection{Eliminating Catastrophic Forgetting (CF)}
\begin{figure*}
	\label{other}
	\centering
	\includegraphics[width=1.0\linewidth]{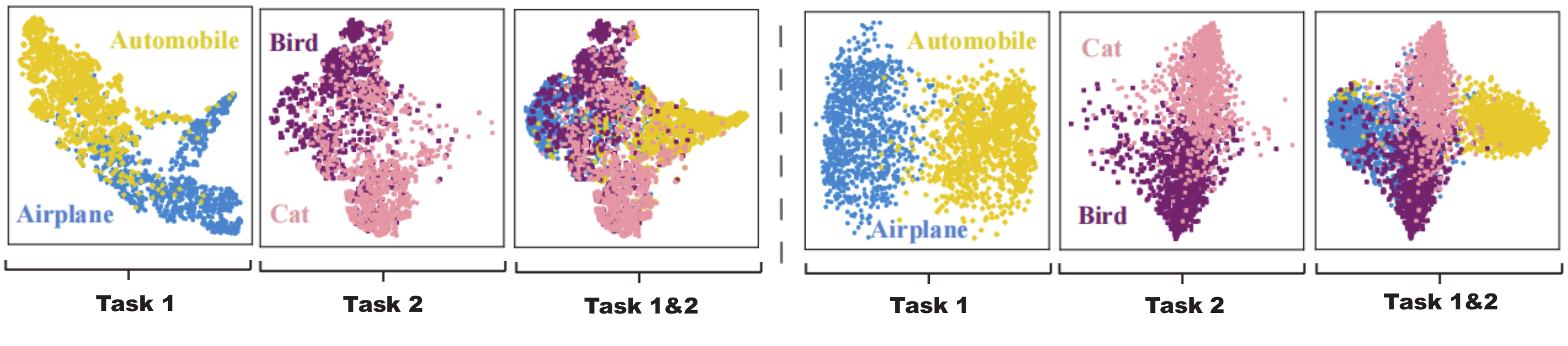}
	\caption{$t$-SNE~\cite{tsne} visualizations of features in ER~\cite{ER} and OnPro~\cite{onpro} on the test set of CIFAR-10. When new classes are coming, the features may overlap with each other, leading to chaos in the embedding space even only two tasks are involved.
	}	
\end{figure*}

\begin{figure*}
	\label{tsne_motivation}
	\centering
	\includegraphics[width=1.0\linewidth]{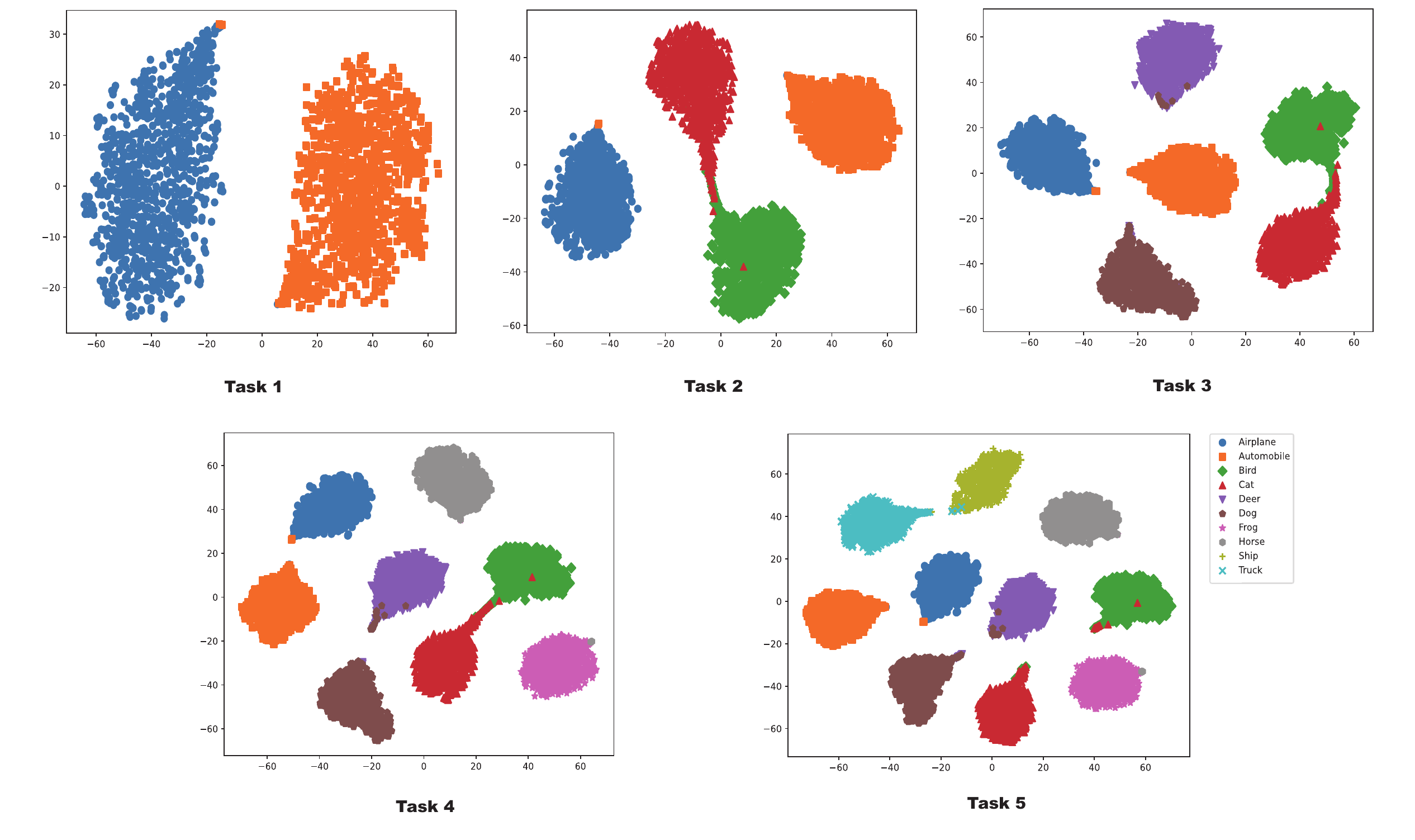}
	\caption{$t$-SNE visualizations of features from the proposed GKDE framework on the test set of CIFAR-10. For each subfigure, such as Task 3, the features are visualized in a unified embedding space where features from previous Task 1 and Task 2 are also included. Despite the sequential addition of the ten classes (Bird, Cat, Deer, Dog, Frog, Horse, Ship, and Truck), they appear to remain separate in the last unified space though the features in each task is coming from different feature space.
	}

\end{figure*}

CF is the primary challenge in online CL as the model struggles to capture sufficient representative features in the embedding feature space due to the single-pass data stream and limited model generalization. To illustrate this issue, we depict some state-of-the-art papers in Fig. 2 and visualize the results using $t$-SNE~\cite{tsne} for ER and OnPro\cite{onpro} on the test set of CIFAR-10. Both methods employ a re-play memory of size ($M=0.2k$) to maximize their capacity. In this setting, there are five tasks, each consisting of two classes. However, we observe severe class confusion in ER and OnPro after task 2 is added, indicating that the features learned in Task 2 are not discriminative enough for Task 1. Consequently, the task confusion leads to a decrease in performance for the previous tasks. In other words, these methods fail to achieve TP and even for the WP their generalization abilities are far from satisfactory. For instance, in Task 2, the Cat and Bird classes in the test set are not well distinguished.

In Fig. 3, our proposed GKDE framework showcases exceptional performance as each task from 1 to 5 is added. In Task 1, the Airplane and Automobile classes are perfectly separated, surpassing the performance of peer classic MLs such as ER and OnPro, as shown in Fig. 2. As new tasks are incrementally carried out, the Bird, Cat, Deer, Dog, Frog, Horse, Ship, and Truck are accommodated sequentially in the feature space without overlapping with each other, the CF issue is well solved. Eventually, all ten classes can be clearly separated in a unified feature space. To the best of our knowledge, no other online CL methods can achieve this level of performance, bridging the gap between online CL and classical ML.

\paragraph{Comparison with State-of-the-art.}

\begin{table*}[ht]
	\small
	\begin{center}
		\resizebox{\linewidth}{!}{
			\begin{tabular}{rrrrrrrrrrrr}
				\shline
				\multirow{2}{*}{Method}  & \multicolumn{3}{c}{CIFAR-10}   && \multicolumn{3}{c}{CIFAR-100}  && \multicolumn{3}{c}{TinyImageNet} \\ \cline{2-4}\cline{6-8}\cline{10-12}
				& $M=0.1k$   & $M=0.2k$   & $M=0.5k$     && $M=0.5k$     & $M=1k$     & $M=2k$     && $M=1k$      & $M=2k$ & $M=4k$   \\ \midrule
				iCaRL~\cite{iCaRL}    & 31.0\std{$\pm$1.2} & 33.9\std{$\pm$0.9} & 42.0\std{$\pm$0.9} && 12.8\std{$\pm$0.4}  & 16.5\std{$\pm$0.4}  & 17.6\std{$\pm$0.5} && 5.0\std{$\pm$0.3}   & 6.6\std{$\pm$0.4} & 7.8\std{$\pm$0.4} \\ 
				DER++~\cite{DER++}   & 31.5\std{$\pm$2.9} & 39.7\std{$\pm$2.7} & 50.9\std{$\pm$1.8} && 16.0\std{$\pm$0.6}  & 21.4\std{$\pm$0.9}  & 23.9\std{$\pm$1.0} && 3.7\std{$\pm$0.4} & 5.1\std{$\pm$0.8} & 6.8\std{$\pm$0.6} \\ 
				PASS~\cite{protoAug}    & 33.7\std{$\pm$2.2} & 33.7\std{$\pm$2.2} & 33.7\std{$\pm$2.2} && 7.5\std{$\pm$0.7}  & 7.5\std{$\pm$0.7}  & 7.5\std{$\pm$0.7} && 0.5\std{$\pm$0.1}   & 0.5\std{$\pm$0.1} & 0.5\std{$\pm$0.1} \\ 
				\hline
				AGEM~\cite{AGEM}    & 17.7\std{$\pm$0.3} & 17.5\std{$\pm$0.3} & 17.5\std{$\pm$0.2} && 5.8\std{$\pm$0.1}  & 5.9\std{$\pm$0.1}  & 5.8\std{$\pm$0.1} && 0.8\std{$\pm$0.1}   & 0.8\std{$\pm$0.1} & 0.8\std{$\pm$0.1} \\ 
				GSS~\cite{GSS}     & 18.4\std{$\pm$0.2} & 19.4\std{$\pm$0.7} & 25.2\std{$\pm$0.9} && 8.1\std{$\pm$0.2}  & 9.4\std{$\pm$0.5}  & 10.1\std{$\pm$0.8} && 1.1\std{$\pm$0.1}   & 1.5\std{$\pm$0.1} & 2.4\std{$\pm$0.4} \\ 
				ER~\cite{ER}      & 19.4\std{$\pm$0.6} & 20.9\std{$\pm$0.9} & 26.0\std{$\pm$1.2} && 8.7\std{$\pm$0.3}  & 9.9\std{$\pm$0.5}  & 10.7\std{$\pm$0.8} && 1.2\std{$\pm$0.1}   & 1.5\std{$\pm$0.2} & 2.0\std{$\pm$0.2} \\ 
				MIR~\cite{MIR}     & 20.7\std{$\pm$0.7} & 23.5\std{$\pm$0.8} & 29.9\std{$\pm$1.2} && 9.7\std{$\pm$0.3}  & 11.2\std{$\pm$0.4}  & 13.0\std{$\pm$0.7} && 1.4\std{$\pm$0.1}   & 1.9\std{$\pm$0.2} & 2.9\std{$\pm$0.3} \\ 
				GDumb~\cite{GDumb}   & 23.3\std{$\pm$1.3} & 27.1\std{$\pm$0.7} & 34.0\std{$\pm$0.8} && 8.2\std{$\pm$0.2}  & 11.0\std{$\pm$0.4}  & 15.3\std{$\pm$0.3} && 4.6\std{$\pm$0.3}   & 6.6\std{$\pm$0.2} & 10.0\std{$\pm$0.3} \\ 
				ASER~\cite{ASER}   & 20.0\std{$\pm$1.0} & 22.8\std{$\pm$0.6} & 31.6\std{$\pm$1.1} && 11.0\std{$\pm$0.3}  & 13.5\std{$\pm$0.3}  & 17.6\std{$\pm$0.4} && 2.2\std{$\pm$0.1}   & 4.2\std{$\pm$0.6} & 8.4\std{$\pm$0.7} \\ 
				SCR~\cite{SCR}     & 40.2\std{$\pm$1.3} & 48.5\std{$\pm$1.5} & 59.1\std{$\pm$1.3} && 19.3\std{$\pm$0.6}  & 26.5\std{$\pm$0.5}  & 32.7\std{$\pm$0.3} && 8.9\std{$\pm$0.3}   & 14.7\std{$\pm$0.3} & 19.5\std{$\pm$0.3} \\ 
				CoPE~\cite{online_pro_ema}  & 33.5\std{$\pm$3.2} & 37.3\std{$\pm$2.2} & 42.9\std{$\pm$3.5} && 11.6\std{$\pm$0.7}  & 14.6\std{$\pm$1.3}  & 16.8\std{$\pm$0.9} && 2.1\std{$\pm$0.3}   & 2.3\std{$\pm$0.4} & 2.5\std{$\pm$0.3} \\
				DVC~\cite{DVC} & 35.2\std{$\pm$1.7}  & 41.6\std{$\pm$2.7} & 53.8\std{$\pm$2.2} &&  15.4\std{$\pm$0.7} & 20.3\std{$\pm$1.0} & 25.2\std{$\pm$1.6} && 4.9\std{$\pm$0.6} &  7.5\std{$\pm$0.5} & 10.9\std{$\pm$1.1} \\ 
				OCM~\cite{OCM} & 47.5\std{$\pm$1.7}  & 59.6\std{$\pm$0.4} & 70.1\std{$\pm$1.5} && 19.7\std{$\pm$0.5} & 27.4\std{$\pm$0.3} & 34.4\std{$\pm$0.5} && 10.8\std{$\pm$0.4} & 15.4\std{$\pm$0.4} & 20.9\std{$\pm$0.7} \\ 
				OnPro~\cite{onpro} & 57.8\std{$\pm$1.1}  & 65.5\std{$\pm$1.0} & 72.6\std{$\pm$0.8} && 22.7\std{$\pm$0.5} & 30.0\std{$\pm$0.3} & 35.9\std{$\pm$0.5} && 11.9\std{$\pm$0.4} & 16.9\std{$\pm$0.4} & 22.1\std{$\pm$0.7} \\ 
				\hline
				\frameworkName (\textbf{ResNet18}) & \multicolumn{3}{c}{90.6\std{$\pm$1.8}} && \multicolumn{3}{c}{78.54\std{$\pm$2.3}} && \multicolumn{3}{c}{82.35\std{$\pm$2.8}}
				\\ 
				\frameworkName (\textbf{VIT}) & \multicolumn{3}{c}{99.25\std{$\pm$0.5}} && \multicolumn{3}{c}{92.88\std{$\pm$1.9}} && \multicolumn{3}{c}{99.10\std{$\pm$0.6}}
				\\ 
				\shline
			\end{tabular}
		}
	\end{center}
	\caption{Average Accuracy~(higher is better) on three benckmark datasets. The baseline methods with different memory bank sizes $M$ exhibit different accuracies. Our method does not need re-play memory bank to store training samples and for each dataset we report one result for ResNet18 and VIT respectively. All results are the average and standard deviation of 15 runs.}
	\label{tab:acc}
\end{table*}

Tab.~\ref{tab:acc} presents the average accuracy results of state-of-the-art methods with different memory bank sizes on three benchmark datasets. Our framework consistently outperforms all of them (re-play or distillation based methods), demonstrating a significant leap that addresses the online CL challenges effectively. For instance our method based on VIT achieves 99.25\% on CIFAR-10 and 99.10\% on TinyImageNet. However, the peer methods exhibit poor performance especially on TinyImageNet due to large number of tasks. Even the GKDE based on ResNet18, a simple backbone, clearly outperforms other methods by a large margin.

\begin{figure}[t]
	\label{incrementalAcc}
	\centering
	\begin{subfigure}[t]{3.0in}
		\centering
		\includegraphics[width=2.8in]{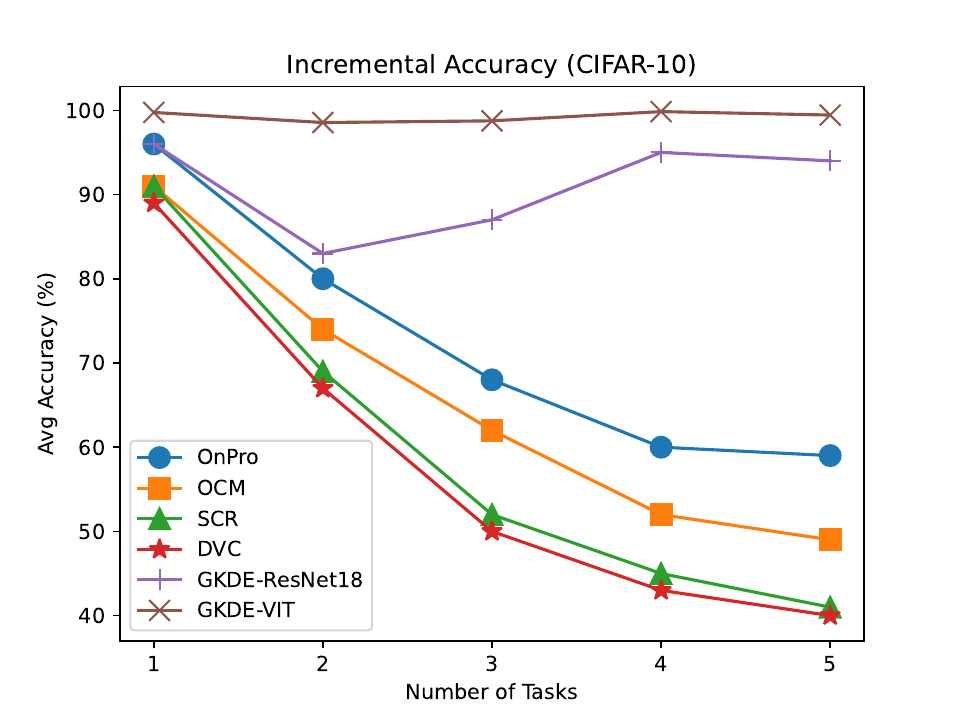}
		\caption{Task incremental average accuracy on CIFAR-10.}
	\end{subfigure}
	\\
	\begin{subfigure}[t]{3.0in}
		\centering
		\includegraphics[width=2.8in]{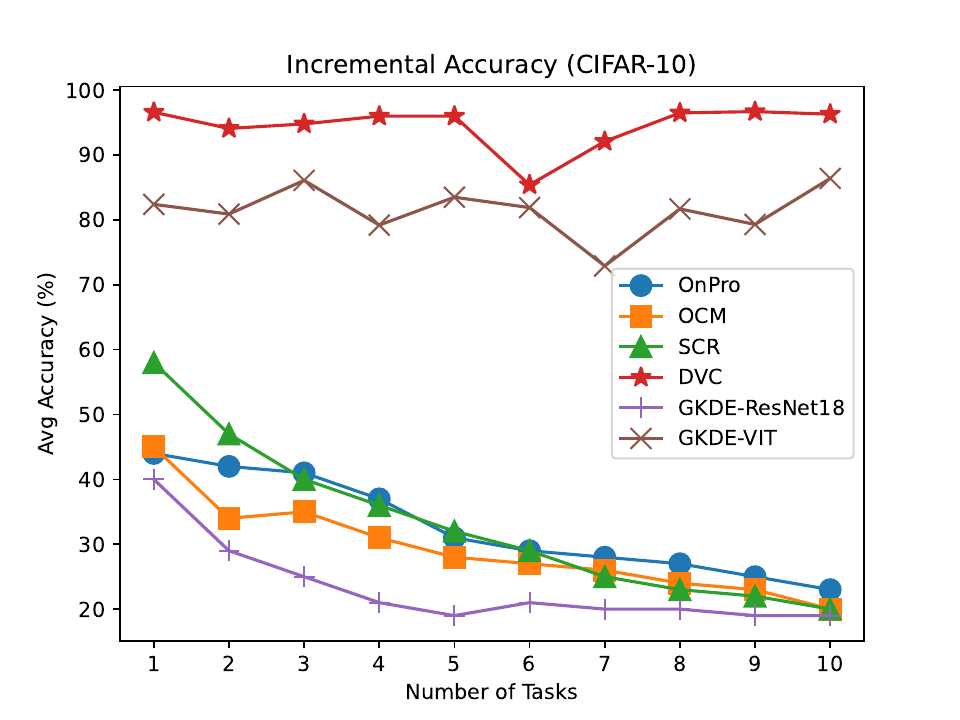}
		\caption{Task incremental average accuracy on CIFAR-100.}
	\end{subfigure}	
	\caption{Average accuracy comparison between the GKDE and peer methods in the test set of CIFAR-10 and CIFAR-100 under the CIL setting. While the other methods are suffering from severe forgetting as new tasks are added, the GKDE based on VIT and ResNet18 achieves excellence performance in the TP and WP procedures as a whole. }
	\label{fig:fourfigures}
\end{figure}

\paragraph{How the performance changes as new tasks are added.}
We compare the average incremental performance of the GKDE framework with peer methods, on CIFAR-10 and CIFAR-100 datasets~\cite{DER++, DVC}. Fig.~4 demonstrates that \frameworkName achieves higher accuracy and effectively solves the issue of forgetting, while the performance of most other methods rapidly deteriorates with the introduction of new classes. The GKDE framework, based on the raw VIT structure, achieves excellent performance in the WP, reporting 99.25\% accuracy on CIFAR-10 and 99.10\% on TinyImageNet. Even the GKDE based on the raw ResNet18 exhibits a clear advantage over other methods. More incremental analysis can be found in Appendix.B. 

\begin{figure}
	\label{bandwidth}
	\centering
	\includegraphics[width=2.9 in,height=2.6 in]{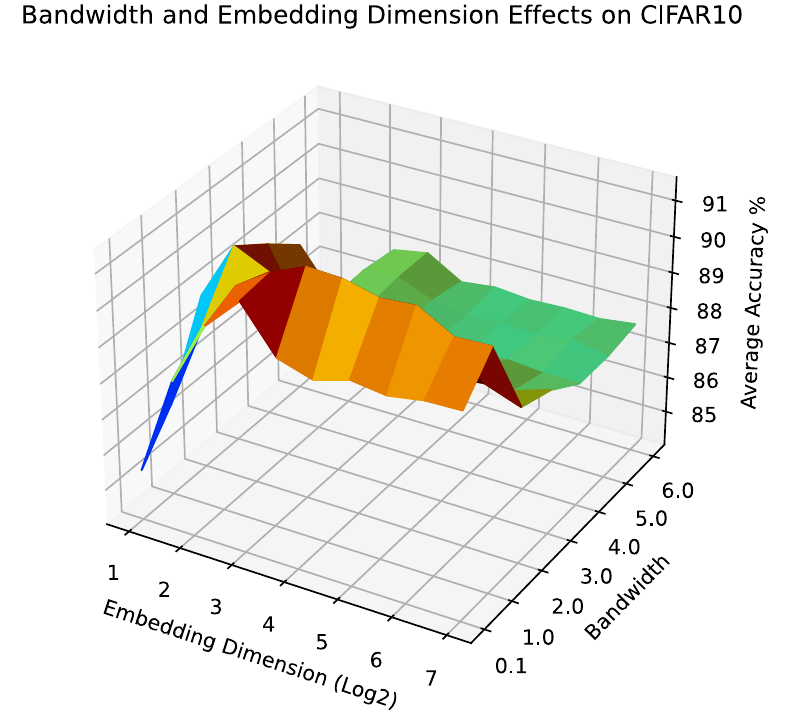}
	\caption{The Bandwidth and Embedding dimension Effect on CIFAR-10. When the dimensions are fewer than 4, it exhibits signs of under-fitting, however there is minimal improvement as the dimension increases to 32 or beyond. When the bandwidth is below 0.1, the performance is unstable and when the bandwidth exceeds 2.0, the framework's performance significantly deteriorates.}
\end{figure}
\paragraph{How the embedding dimension and bandwidth affect the performance.}
The embedding dimension and bandwidth are crucial factors in determining the GKDE learning objective. In this section, we validate how these parameters affect the performance of our framework. To identify the optimal parameter combination, we conducted a grid search over the embedding dimensions ranging from 2 to 128 (2, 4, 8, 16, 32, 64, 128) and bandwidth values ranging from 0.1 to 6 (0.1, 1, 2, 3, 4, 5, 6), as illustrated in Fig. 5. When the embedding dimensions are fewer than 4, the framework exhibits signs of under-fitting, and there is minimal improvement as the dimension increases to 32 or beyond. On the other hand, when the bandwidth is set below 0.1, the performance becomes unstable and leads to lower accuracy and when it exceeds 2.0, the framework's performance significantly deteriorates. Based on the results of the grid search, we have selected an embedding dimension of 32 and a bandwidth of 0.5 for CIFAR-10. For CIFAR-100 and TinyImageNet, we have chosen an embedding dimension of 64 and a bandwidth of 0.3.
\paragraph{How the sample size affect model performance and time-space efficiency.}
For each class in a task, the GKDE samples a certain number of embedding features. In order to assess the impact of the sample size on model performance and time-space efficiency, we experimented with various values ranging from 100 to 3000. Our findings indicate that a sample size of 500 yields the best results for CIFAR-10, while for CIFAR-100 and TinyImageNet, a sample size of 1500 is the optimal choice. Interestingly, increasing the sample size beyond these values does not lead to further performance improvements. Instead, it negatively affects the time-space efficiency during both the training and inference stages. For the memory usage in CIFAR-10, our GKDE requires only one quarter of the memory compared to replay-based methods. Specifically, we need 500 * 32 bytes for each class, while the replay-based methods require a minimum of 200 * 32 * 32 * 3 bytes to store the original images. Regarding training efficiency, as shown in Fig. 6, we compared the GKDE with peer methods. Our method completes 5 tasks with 10 epochs per task in approximately 1 minute and each epoch runs in just 1.5 seconds on CIFAR-10 with one GTX 4090. More analysis, limitations can be found in Appendix.C.

\begin{figure}
	\label{efficieny}
	\centering
	\includegraphics[width=3.1 in,height=2.8 in]{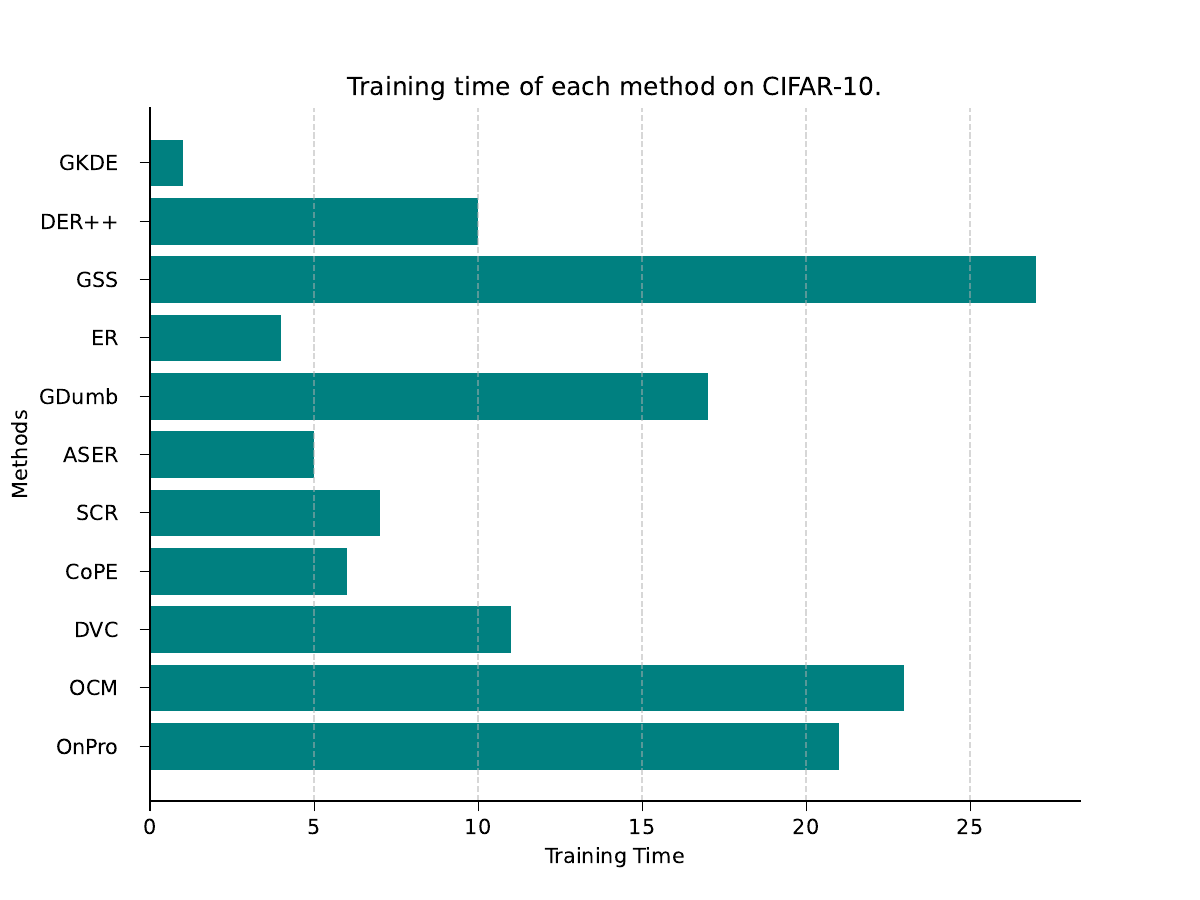}
	\caption{Training Time Efficiency Comparison (In Minutes). Our method completes 5 tasks with 10 epochs per task in approximately 1 minute and each epoch runs in just 1.5 seconds on CIFAR-10. The training time is much shorter than other methods. }
\end{figure}
\section{Conclusion}
In this paper, we introduce a density distribution-based learning framework for addressing the challenges of online CL. Our method strategically groups samples with the same label into high-density regions, while simultaneously pushing dissimilar samples apart. Thanks to MB structure, the method tackles the TP and WP simultaneously. This approach leads to superior average accuracy, outperforming existing peer methods by a substantial leap and effectively addressing the DF issue. The GKDE generalization is validated on several datasets with improved feature representation. Additionally, our proposed methods demonstrate commendable time and space efficiency, making our method more suitable for real-world ML applications.

{
    \small
    \bibliographystyle{ieeenat_fullname}
    \bibliography{main}

\begin{thebibliography}{45}
\providecommand{\natexlab}[1]{#1}
\providecommand{\url}[1]{\texttt{#1}}
\expandafter\ifx\csname urlstyle\endcsname\relax
  \providecommand{\doi}[1]{doi: #1}\else
  \providecommand{\doi}{doi: \begingroup \urlstyle{rm}\Url}\fi

\bibitem[Abati et~al.(2020)Abati, Tomczak, Blankevoort, Calderara, Cucchiara,
  and Bejnordi]{abati2020conditional}
Davide Abati, Jakub Tomczak, Tijmen Blankevoort, Simone Calderara, Rita
  Cucchiara, and Ehteshami Bejnordi.
\newblock Conditional channel gated networks for task-aware continual learning.
\newblock In \emph{CVPR}, pages 3931--3940, 2020.

\bibitem[Aljundi et~al.(2019{\natexlab{a}})Aljundi, Belilovsky, Tuytelaars,
  Charlin, Caccia, and Lin.Min.and Page-Caccia]{MIR}
Rahaf Aljundi, Eugene Belilovsky, Tinne Tuytelaars, Laurent Charlin, Massimo
  Caccia, and Lucas Lin.Min.and Page-Caccia.
\newblock Online continual learning with maximal interfered retrieval.
\newblock In \emph{Advances in Neural Information Processing Systems},
  2019{\natexlab{a}}.

\bibitem[Aljundi et~al.(2019{\natexlab{b}})Aljundi, Lin, Goujaud, and
  Bengio]{GSS}
Rahaf Aljundi, Min Lin, Baptiste Goujaud, and Yoshua Bengio.
\newblock Gradient based sample selection for online continual learning.
\newblock \emph{Advances in Neural Information Processing Systems}, 32,
  2019{\natexlab{b}}.

\bibitem[Buzzega et~al.(2020)Buzzega, Boschini, Porrello, Abati, and
  Calderara]{DER++}
Pietro Buzzega, Matteo Boschini, Angelo Porrello, Davide Abati, and Simone
  Calderara.
\newblock Dark experience for general continual learning: a strong, simple
  baseline.
\newblock \emph{Advances in Neural Information Processing Systems},
  33:\penalty0 15920--15930, 2020.

\bibitem[Caccia et~al.(2022)Caccia, Aljundi, Asadi, Tuytelaars, Pineau, and
  Belilovsky]{ER_AML}
Lucas Caccia, Rahaf Aljundi, Nader Asadi, Tinne Tuytelaars, Joelle Pineau, and
  Eugene Belilovsky.
\newblock New insights on reducing abrupt representation change in online
  continual learning.
\newblock \emph{arXiv:2203.03798}, 2022.

\bibitem[Cha et~al.(2021)Cha, Lee, and Shin]{Co2L}
Hyuntak Cha, Jaeho Lee, and Jinwoo Shin.
\newblock Co\({}^{\mbox{2}}\)l: Contrastive continual learning.
\newblock In \emph{Proceedings of the IEEE/CVF International Conference on
  Computer Vision}, pages 9516--9525, 2021.

\bibitem[Chaudhry et~al.(2018)Chaudhry, Ranzato, Rohrbach, and Elhoseiny]{AGEM}
Arslan Chaudhry, Marc'Aurelio Ranzato, Marcus Rohrbach, and Mohamed Elhoseiny.
\newblock Efficient lifelong learning with {A-GEM}.
\newblock \emph{arXiv:1812.00420}, 2018.

\bibitem[Chaudhry et~al.(2019)Chaudhry, Rohrbach, Elhoseiny, Ajanthan, Dokania,
  Torr, and Ranzato]{ER}
Arslan Chaudhry, Marcus Rohrbach, Mohamed Elhoseiny, Thalaiyasingam Ajanthan,
  Puneet~K Dokania, Philip~HS Torr, and Marc'Aurelio Ranzato.
\newblock On tiny episodic memories in continual learning.
\newblock \emph{arXiv:1902.10486}, 2019.

\bibitem[Chrysakis and Moens(2020)]{onlineCL1}
Aristotelis Chrysakis and Marie-Francine Moens.
\newblock Online continual learning from imbalanced data.
\newblock In \emph{International Conference on Machine Learning}, pages
  1952--1961, 2020.

\bibitem[der Maaten and Hinton(2008)]{tsne}
Laurens~Van der Maaten and Geoffrey Hinton.
\newblock Visualizing data using t-sne.
\newblock \emph{Journal of Machine Learning Research}, 9\penalty0 (11), 2008.

\bibitem[Dosovitskiy et~al.(2020)Dosovitskiy, Beyer, Kolesnikov, Weissenborn,
  Zhai, Unterthiner, Dehghani, Minderer, Heigold, Gelly, et~al.]{ViT}
Alexey Dosovitskiy, Lucas Beyer, Alexander Kolesnikov, Dirk Weissenborn,
  Xiaohua Zhai, Thomas Unterthiner, Mostafa Dehghani, Matthias Minderer, Georg
  Heigold, Sylvain Gelly, et~al.
\newblock An image is worth 16x16 words: Transformers for image recognition at
  scale.
\newblock \emph{arXiv:2010.11929}, 2020.

\bibitem[Fernando et~al.(2017)Fernando, Banarse, Blundell, Zwols, Ha, Rusu,
  Pritzel, and Wierstra]{pathNet}
Chrisantha Fernando, Dylan Banarse, Charles Blundell, Yori Zwols, David Ha,
  Andrei~A. Rusu, Alexander Pritzel, and Daan Wierstra.
\newblock {PathNet}: Evolution channels gradient descent in super neural
  networks.
\newblock \emph{arXiv:1701.08734}, 2017.

\bibitem[Fini et~al.(2022)Fini, Costa, Alameda-Pineda, Ricci, Alahari, and
  Mairal]{CL_CIL1}
Enrico Fini, Victor G. Turrisi~Da Costa, Xavier Alameda-Pineda, Elisa Ricci,
  Karteek Alahari, and Julien Mairal.
\newblock Self-supervised models are continual learners.
\newblock In \emph{Proceedings of the IEEE/CVF Conference on Computer Vision
  and Pattern Recognition}, pages 9621--9630, 2022.

\bibitem[French(1999)]{catastrophic}
Robert~M French.
\newblock Catastrophic forgetting in connectionist networks.
\newblock \emph{Trends in Cognitive Sciences}, 3\penalty0 (4):\penalty0
  128--135, 1999.

\bibitem[Goodfellow et~al.(2013)Goodfellow, Mirza, Xiao, Courville, and
  Bengio]{catastrophic2}
Ian~J Goodfellow, Mehdi Mirza, Da Xiao, Aaron Courville, and Yoshua Bengio.
\newblock An empirical investigation of catastrophic forgetting in
  gradient-based neural networks.
\newblock \emph{arXiv:1312.6211}, 2013.

\bibitem[Gu et~al.(2022)Gu, Yang, Wei, and Deng]{DVC}
Yanan Gu, Xu Yang, Kun Wei, and Cheng Deng.
\newblock Not just selection, but exploration: Online class-incremental
  continual learning via dual view consistency.
\newblock In \emph{Proceedings of the IEEE/CVF Conference on Computer Vision
  and Pattern Recognition}, pages 7442--7451, 2022.

\bibitem[Guo et~al.(2022)Guo, Liu, and Zhao]{OCM}
Yiduo Guo, Bing Liu, and Dongyan Zhao.
\newblock Online continual learning through mutual information maximization.
\newblock \emph{International Conference on Machine Learning}, pages
  8109--8126, 2022.

\bibitem[He and Zhu(2022)]{online_pro_accum}
Jiangpeng He and Fengqing Zhu.
\newblock Exemplar-free online continual learning.
\newblock In \emph{2022 IEEE International Conference on Image Processing
  (ICIP)}, pages 541--545, 2022.

\bibitem[He et~al.(2020)He, Mao, Shao, and Zhu]{onlineCL2}
Jiangpeng He, Runyu Mao, Zeman Shao, and Fengqing Zhu.
\newblock Incremental learning in online scenario.
\newblock In \emph{Proceedings of the IEEE/CVF Conference on Computer Vision
  and Pattern Recognition}, pages 13926--13935, 2020.

\bibitem[He et~al.(2016)He, Zhang, Ren, and Sun]{ResNet}
Kaiming He, Xiangyu Zhang, Shaoqing Ren, and Jian Sun.
\newblock Deep residual learning for image recognition.
\newblock In \emph{Proceedings of the IEEE Conference on Computer Vision and
  Pattern Recognition}, pages 770--778, 2016.

\bibitem[Henning et~al.(2021)Henning, Cervera, D'Angelo, Von~Oswald, Traber,
  Ehret, Kobayashi, Grewe, and Sacramento]{henning2021posterior}
Christian Henning, Maria Cervera, Francesco D'Angelo, Johannes Von~Oswald,
  Regina Traber, Benjamin Ehret, Seijin Kobayashi, Benjamin~F Grewe, and
  Jo{\~a}o Sacramento.
\newblock Posterior meta-replay for continual learning.
\newblock \emph{NeurIPS}, 34, 2021.

\bibitem[Hung et~al.(2019)Hung, Tu, Wu, Chen, Chan, and Chen]{hung2019compact}
Ching-Yi Hung, Cheng-Hao Tu, Cheng-En Wu, Chien-Hung Chen, Yi-Ming Chan, and
  Chu-Song Chen.
\newblock Compacting, picking and growing for unforgetting continual learning.
\newblock In \emph{NeurIPS}, 2019.

\bibitem[Krizhevsky et~al.(2009)Krizhevsky, Hinton, and et~al.]{cifar10_100}
Alex Krizhevsky, Geoffrey Hinton, and et al.
\newblock Learning multiple layers of features from tiny images.
\newblock \emph{Technical Report TR-2009, University of Toronto, Toronto},
  2009.

\bibitem[Lange and Tuytelaars(2021)]{online_pro_ema}
Matthias~De Lange and Tinne Tuytelaars.
\newblock Continual prototype evolution: Learning online from non-stationary
  data streams.
\newblock In \emph{Proceedings of the IEEE/CVF International Conference on
  Computer Vision}, pages 8250--8259, 2021.

\bibitem[Lange et~al.(2021)Lange, Aljundi, Masana, Parisot, Jia,
  et~al.]{survey3}
Matthias~De Lange, Rahaf Aljundi, Marc Masana, Sarah Parisot, Xu Jia, et~al.
\newblock A continual learning survey: Defying forgetting in classification
  tasks.
\newblock \emph{IEEE Transactions on Pattern Analysis and Machine
  Intelligence}, 44\penalty0 (7):\penalty0 3366--3385, 2021.

\bibitem[Le and Yang(2015)]{tinyImageNet}
Ya Le and Xuan Yang.
\newblock Tiny {ImageNet} visual recognition challenge.
\newblock \emph{CS 231N}, 7\penalty0 (7):\penalty0 3, 2015.

\bibitem[Li et~al.(2022)Li, Dai, Ge, Liu, Shan, and DUAN]{li2022uncertainty}
Xiaotong Li, Yongxing Dai, Yixiao Ge, Jun Liu, Ying Shan, and LINGYU DUAN.
\newblock Uncertainty modeling for out-of-distribution generalization.
\newblock In \emph{International Conference on Learning Representations}, 2022.

\bibitem[Mai et~al.(2021)Mai, Li, Kim, and Sanner]{SCR}
Zheda Mai, Ruiwen Li, Hyunwoo Kim, and Scott Sanner.
\newblock Supervised contrastive replay: Revisiting the nearest class mean
  classifier in online class-incremental continual learning.
\newblock In \emph{Proceedings of the IEEE/CVF Conference on Computer Vision
  and Pattern Recognition (CVPR) Workshops}, pages 3589--3599, 2021.

\bibitem[Mai et~al.(2022)Mai, Li, Jeong, Quispe, Kim, and
  Sanner]{online_survey}
Zheda Mai, Ruiwen Li, Jihwan Jeong, David Quispe, Hyunwoo Kim, and Scott
  Sanner.
\newblock Online continual learning in image classification: An empirical
  survey.
\newblock \emph{Neurocomputing}, 469:\penalty0 28--51, 2022.

\bibitem[Mallya and Lazebnik(2017)]{Mallya2017packnet}
Arun Mallya and Svetlana Lazebnik.
\newblock {PackNet: Adding Multiple Tasks to a Single Network by Iterative
  Pruning}.
\newblock \emph{arXiv preprint arXiv:1711.05769}, 2017.

\bibitem[Masana et~al.(2022)Masana, Liu, Twardowski, Menta, Bagdanov, and
  van~de Weijer]{survey2}
Marc Masana, Xialei Liu, Bartłomiej Twardowski, Mikel Menta, Andrew~D
  Bagdanov, and Joost van~de Weijer.
\newblock Class-incremental learning: survey and performance evaluation on
  image classification.
\newblock \emph{IEEE Transactions on Pattern Analysis and Machine
  Intelligence}, 2022.

\bibitem[Parisi et~al.(2019)Parisi, Kemker, Part, Kanan, and Wermter]{survey1}
German~I Parisi, Ronald Kemker, Jose~L Part, Christopher Kanan, and Stefan
  Wermter.
\newblock Continual lifelong learning with neural networks: A review.
\newblock \emph{Neural networks}, 113:\penalty0 54--71, 2019.

\bibitem[Prabhu et~al.(2020)Prabhu, Torr, and Dokania]{GDumb}
Ameya Prabhu, Philip~HS Torr, and Puneet~K Dokania.
\newblock Gdumb: A simple approach that questions our progress in continual
  learning.
\newblock In \emph{Proceedings of the European Conference on Computer Vision
  (ECCV)}, pages 524--540, 2020.

\bibitem[Rajasegaran et~al.(2020)Rajasegaran, Khan, Hayat, Khan, and
  Shah]{rajasegaran2020itaml}
Jathushan Rajasegaran, Salman Khan, Munawar Hayat, Fahad~Shahbaz Khan, and
  Mubarak Shah.
\newblock itaml: An incremental task-agnostic meta-learning approach.
\newblock In \emph{CVPR}, 2020.

\bibitem[Rebuffi et~al.(2017)Rebuffi, Kolesnikov, Sperl, and Lampert]{iCaRL}
Sylvestre-Alvise Rebuffi, Alexander Kolesnikov, Georg Sperl, and Christoph~H
  Lampert.
\newblock icarl: Incremental classifier and representation learning.
\newblock In \emph{Proceedings of the IEEE conference on Computer Vision and
  Pattern Recognition}, pages 2001--2010, 2017.

\bibitem[Rusu et~al.(2016)Rusu, Rabinowitz, Desjardins, Soyer, Kirkpatrick,
  Kavukcuoglu, Pascanu, and Hadsell]{PNN}
Andrei~A Rusu, Neil~C Rabinowitz, Guillaume Desjardins, Hubert Soyer, James
  Kirkpatrick, Koray Kavukcuoglu, Razvan Pascanu, and Raia Hadsell.
\newblock Progressive neural networks.
\newblock \emph{arXiv:1606.04671}, 2016.

\bibitem[Serr{\`{a}} et~al.(2018)Serr{\`{a}}, Sur{\'{i}}s, Miron, and
  Karatzoglou]{Serra2018overcoming}
Joan Serr{\`{a}}, D{\'{i}}dac Sur{\'{i}}s, Marius Miron, and Alexandros
  Karatzoglou.
\newblock {Overcoming catastrophic forgetting with hard attention to the task}.
\newblock In \emph{ICML}, 2018.

\bibitem[Serra et~al.(2018)Serra, Suris, Miron, and Karatzoglou]{para-iso1}
Joan Serra, Didac Suris, Marius Miron, and Alexandros Karatzoglou.
\newblock Overcoming catastrophic forgetting with hard attention to the task.
\newblock In \emph{International Conference on Machine Learning}, pages
  4548--4557, 2018.

\bibitem[Shim et~al.(2021)Shim, Mai, Jeong, Sanner, Kim, and Jang]{ASER}
Dongsub Shim, Zheda Mai, Jihwan Jeong, Scott Sanner, Hyunwoo Kim, and Jongseong
  Jang.
\newblock Online class-incremental continual learning with adversarial shapley
  value.
\newblock In \emph{Proceedings of the AAAI Conference on Artificial
  Intelligence}, pages 9630--9638, 2021.

\bibitem[von Oswald et~al.(2020)von Oswald, Henning, Sacramento, and
  Grewe]{von2019continual_hypernet}
Johannes von Oswald, Christian Henning, Jo{\~a}o Sacramento, and Benjamin~F
  Grewe.
\newblock Continual learning with hypernetworks.
\newblock \emph{ICLR}, 2020.

\bibitem[Wei et~al.(2023)Wei, Ye, Huang, Zhang, and Shan]{onpro}
Yujie Wei, Jiaxin Ye, Zhizhong Huang, Junping Zhang, and Hongming Shan.
\newblock Online prototype learning for online continual learning.
\newblock In \emph{ICCV}, 2023.

\bibitem[Wortsman et~al.(2020)Wortsman, Ramanujan, Liu, Kembhavi, Rastegari,
  Yosinski, and Farhadi]{supsup2020}
Mitchell Wortsman, Vivek Ramanujan, Rosanne Liu, Aniruddha Kembhavi, Mohammad
  Rastegari, Jason Yosinski, and Ali Farhadi.
\newblock Supermasks in superposition.
\newblock In \emph{NeurIPS}, 2020.

\bibitem[Zhang et~al.(2021{\natexlab{a}})Zhang, Zhang, Zhang, Jin, Zhou, Cai,
  Zhao, Yi, Liu, and Liu]{transformer}
Chongzhi Zhang, Mingyuan Zhang, Shanghang Zhang, Daisheng Jin, Qiang Zhou,
  Zhongang Cai, Haiyu Zhao, Shuai Yi, Xianglong Liu, and Ziwei Liu.
\newblock Delving deep into the generalization of vision transformers under
  distribution shifts.
\newblock \emph{CoRR}, abs/2106.07617, 2021{\natexlab{a}}.

\bibitem[Zhang et~al.(2021{\natexlab{b}})Zhang, Cui, Xu, Zhou, He, and
  Shen]{stable_learning}
Xingxuan Zhang, Peng Cui, Renzhe Xu, Linjun Zhou, Yue He, and Zheyan Shen.
\newblock Deep stable learning for out-of-distribution generalization.
\newblock In \emph{2021 IEEE/CVF Conference on Computer Vision and Pattern
  Recognition (CVPR)}, pages 5368--5378, 2021{\natexlab{b}}.

\bibitem[Zhu et~al.(2021)Zhu, Zhang, Wang, Yin, and Liu]{protoAug}
Fei Zhu, Xu-Yao Zhang, Chuang Wang, Fei Yin, and Cheng-Lin Liu.
\newblock Prototype augmentation and self-supervision for incremental learning.
\newblock In \emph{Proceedings of the IEEE/CVF Conference on Computer Vision
  and Pattern Recognition}, pages 5871--5880, 2021.

\end{thebibliography}
}
\end{document}